\documentclass[conference]{IEEEtran}
\IEEEoverridecommandlockouts
\usepackage{cite}
\usepackage{amsmath,amssymb,amsfonts}
\usepackage{algorithmic}
\usepackage{graphicx}
\usepackage{textcomp}
\usepackage{xcolor}
\usepackage{dblfloatfix}

\def\BibTeX{{\rm B\kern-.05em{\sc i\kern-.025em b}\kern-.08em
    T\kern-.1667em\lower.7ex\hbox{E}\kern-.125emX}}
\begin{document}

\title{Improving English to Sinhala Neural Machine Translation using Part-of-Speech Tag\\
}

\author{\IEEEauthorblockN{Ravinga Perera, Thilakshi Fonseka, Rashmini Naranpanawa, Uthayasanker Thayasivam}
\IEEEauthorblockA{\textit{Department of Computer Science and Engineering } \\
\textit{University of Moratuwa}\\
Katubedda 10400, Sri Lanka \\
\{ravinga.16, thilakshifonseka.16, rashmini.16, rtuthaya\}@cse.mrt.ac.lk}
}

\maketitle

\begin{abstract}
The performance of Neural Machine Translation (NMT) depends significantly on the size of the available parallel corpus. Due to this fact, low resource language pairs demonstrate low translation performance compared to high resource language pairs. The translation quality further degrades when NMT is performed for morphologically rich languages. Even though the web contains a large amount of information, most people in Sri Lanka are unable to read and understand English properly. Therefore, there is a huge requirement of translating English content to local languages to share information among locals. Sinhala language is the primary language in Sri Lanka and building an NMT system that can produce quality English to Sinhala translations is difficult due to the syntactic divergence between these two languages under low resource constraints. Thus, in this research, we explore effective methods of incorporating Part of Speech (POS) tags to the Transformer input embedding and positional encoding to further enhance the performance of the baseline English to Sinhala neural machine translation model.
\end{abstract}


\section{Introduction}
Neural Machine Translation (NMT) has emerged to be a technique that produces promising results due to the invention of different models. This technique requires a large corpus to demonstrate quality translations compared to Statistical Machine Translation (SMT) [1]. Therefore, the translation quality of low resource language pairs is lagging behind the published high resource NMT benchmarks [2]. 

English-Sinhala language pair falls under the low resource category due to the limited number of available resources. These two languages do not follow the same grammatical structure which also negatively impact the NMT performance. English is a West Germanic language with a word order of Subject-Verb-Object (SVO). However, the Sinhala language belongs to the Indo-Aryan language family which is completely different from English by its Subject-Object-Verb (SOV) word order, morphological variation and grammatical structure. For example, a Sinhala verb is varied due to gender, singular/plural nature etc. thus exhibiting its morphological richness.  

Due to these reasons, it’s immensely challenging to build an effective NMT system that can produce quality translations  which are acceptable. Therefore, various advanced techniques such as syntactic structures specific to a language must be incorporated to mitigate the problems caused when translating between distant languages. In this research, we propose an English to Sinhala NMT system based on Transformer [3] using Part of Speech (POS) information focusing on the parallel corpus built with Sri Lankan official government documents.

Part of Speech (POS) is a label allocated to each word in a particular sentence. POS tags help to disambiguate between nouns, adjectives, verbs, adverbs etc. For example, the word “book” can be a verb or a noun in a particular sentence based on the context. Thus the model must be capable of identifying the correct interpretation of the word when translating it to Sinhala. Further, due to the word order divergence between the two languages, the model must learn which English word must be translated first in the source sentence. Existing literature has proved the advantages of incorporating linguistic features such as POS information in state-of-the-art NMT models [4]. However, most of them are tested on resourceful languages. Only a few published literature is available for under-resourced languages. Due to these facts, we were motivated to explore two ways of incorporating POS tags into the Transformer NMT framework, input embedding and positional encoding. Further, we empirically experiment on integrating POS information to the encoder side by preserving the original Transformer architecture in order to preserve its efficiency. 

The sections of this paper are organized as follows. Section 2 and Section 3 illustrate the background/related-work and methodology of our English to Sinhala translator respectively. In section 4, we describe the experiments we conducted and the results are reported in section 5. Our experiments demonstrate that our Part of Speech (POS) integrated NMT model outperforms the existing English to Sinhala translation benchmark value even with a handful of parallel corpus. Finally, we conclude our research in section 6.

\section{Background and related work
}

\subsection{Neural Machine Translation}

Neural Machine Translation (NMT) is a machine translation technique that is built upon artificial neural networks to facilitate rapid translation and enhance the performance of translation quality. Sequence to sequence mapping was first addressed by the Recurrent Neural Network that also laid the foundation for neural machine translation. Later on, more advanced sequence mapping architectures such as Long Short-Term Memory (LSTM) [5] were introduced to address the problem that occurred with long-range dependencies. The main issue of those architectures was the one-to-one correspondence between the input words and the output words. To address this, the encoder-decoder architecture [6] was proposed, which is the first successful NMT architecture that was developed to address the machine translation problem. However, the main restriction was the need for capturing the input sequence to a fixed-length vector. Thus as a solution, Attention [7] mechanism came into the literature. It improves the translation quality by targeting the relevant sections in the input sequence.

All the architectures mentioned previously were built upon the sequential nature. Therefore, a robust architecture that incorporates parallelization named Transformer [3] was introduced to speed up the training. The Transformer itself is a variant of encoder-decoder architecture with two new techniques named self-attention and positional encoding. Transformer encoder component consists of multiple encoders and the decoder component consists of the same number of decoders. The self-attention layer assists the encoder to focus the other words in the sentence, while it encodes a specific word. The positional encoding was introduced to represent the order of the words in the input sentence. The Transformer is the state-of-the-art NMT architecture which has indeed shown better results for high resource language pairs, but the performance of low resource language pairs is yet arguable. Thus, we were motivated to address English to Sinhala translation using the Transformer architecture.

\subsection{English to Sinhala Neural Machine Translation}

Improving the performance of low-resource NMT systems is difficult since neural methods require an abundance of data for training. Therefore, there is only a few published research available on the English to Sinhala NMT task.

Sen et al. [8] have implemented two separate multilingual NMT models using the Transformer architecture. One model translates English into seven other Indic languages. The other model translates seven Indic languages into English. These seven Indic languages include Sinhala. They have implemented fourteen baseline bilingual models in order to evaluate their multilingual models and the English to Sinhala bilingual model has achieved a BLEU  score of 12.75.

Guzman et al. [9] have implemented an NMT system with the Transformer architecture [3] to introduce a dataset for evaluating English-Sinhala language pair. Their system is an open-domain NMT system. They have carried out training under supervised, unsupervised, semi-supervised and weakly supervised training settings. For the English to Sinhala translation task, they have achieved a BLEU score of 1.2 under the supervised training settings and they have stated this lower BLEU score is due to the domain mismatch between test and training sets. 

The above approach was further improved by Nguyen et al. [10]. They have used a method named data diversification for their implementation. Using two different models named backward peer model and forward peer model, they have improved the size of the available dataset and achieved a BLEU score of 2.2 for the English to Sinhala translation task. This BLEU score surpassed the BLEU score produced by Guzman et al. [9].

Fonseka et al. [11] have built an English to Sinhala NMT model using the Transformer architecture along with Byte Pair Encoding (BPE). Authors have employed a closed domain dataset with 18k parallel sentences, created using Sri Lankan official government documents. They have achieved a BLEU score of 28.28 for the English to Sinhala translation task which is a high BLEU score compared to aforementioned published scores.

Naranpanawa et al. [12] have explored various subword techniques to identify the best suited technique for the English to Sinhala NMT task. Specifically, they have experimented with Byte Pair Encoding (BPE), Unigram Language Model, Subword Sampling, BPE-dropout, and finally Character Segmentation. They have used the Transformer architecture and shown NMT along with subword segmentation strategies improve the translation quality when translating from English to morphologically rich languages. A corpus with 54k parallel sentences built using Sri Lankan official government documents have been used and the authors have obtained the best BLEU score of 29.92 for the model incorporated with BPE.

\subsection{Part of Speech Tagging}

Sennrich et al. [13] have introduced the first NMT model where the translation quality is improved using linguistic features. The authors have used English and German languages for the bilingual neural machine translation. They have proposed an attentional encoder-decoder architecture which allows several input features such as  morphological features, part-of-speech tags and syntactic dependency labels at the encoder. The paper proves that adding linguistic features to the source language improves the translation quality.

Recent research [4] has explored effective ways to improve neural machine translation by adding syntax into the Transformer model in terms of input embedding and positional encoding. They have experimented with various syntactic features such as dependency label, POS tag, parent position and tree depth of the node, considering German-English language pair. Their experimental results have shown that the performance of the Transformer can be improved using syntactic information. 

Hailiang et al. [14] have implemented an improved Transformer architecture for the English-Chinese language translation. They have proposed an improved sinusoidal positional encoding method and it has been further improved by incorporating additional linguistic knowledge such as part-of-speech tagging. They have concluded that their proposed models have achieved improvements compared to the vanilla Transformer.

However, there is only few research on machine translation with linguistic knowledge for low-resource languages. A phrase-based statistical machine translation model has been implemented by incorporating various preprocessing techniques such as POS integration, generating phrasal units and segmentation for Sinhala-Tamil language pair [15]. They have outperformed the baseline results with POS integration and segmentation. Overall, the paper concludes that incorporating preprocessing techniques can further improve the baseline results.

Even though there are several published research on English-Sinhala language translation using Transformer architecture, none of those has integrated any linguistic features. The literature has stated that integrating linguistic knowledge into the NMT system enhances the performance further. However, none of those research has been conducted on very low resource Asian languages. In our research, both source and target languages are low in count.  Therefore, we experiment with existing techniques of POS integration to find out the best technique for the English-Sinhala language pair.

\section{Methodology}
Even though the Transformer architecture [3] is the current state-of-the-art NMT architecture, its performance can be improved by incorporating additional linguistic information into the system as stated in the literature [4][14]. In this research, we built a domain-specific Transformer based system incorporating part-of-speech information as an additional linguistic feature in the source. We used the English-Sinhala BPE model implemented by Naranpanawa et al. [12] as our benchmark model to compare with the augmented POS implementations. 

Our proposed model contains two components to incorporate part-of-speech information, a POS tagging component to assign POS tags and an augmented Transformer model which integrates POS information into the input embedding and positional encoding. The following subsections describe these two components comprehensively.

\subsection{Part of Speech Tagging }\label{AA}
In our research, we incorporated Part-of-Speech (POS) information in the source-side. A POS tag is assigned to each word considering the context of the word in a sentence. Since NMT heavily relies on subword segmentation to overcome the out of vocabulary problem, we used BPE [16] as the subword algorithm. Therefore, we had to consider the mismatch between the words and the subwords when assigning POS tags. Hence, all the subwords of a particular word have been assigned the same POS tag of that particular word.

\begin{figure}[t]
\centerline{\includegraphics[width=8cm, height=8cm]{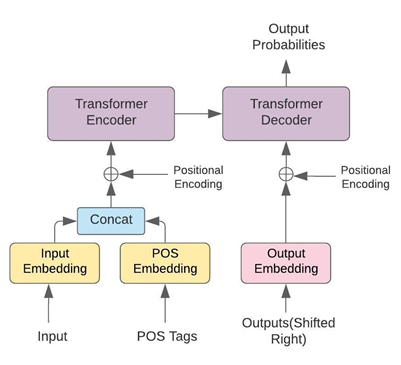}}
\caption{Transformer Architecture of POS Integration into the Input Embedding}
\label{fig}
\end{figure}

\begin{figure}[t]
\centerline{\includegraphics[width=9cm, height=8cm]{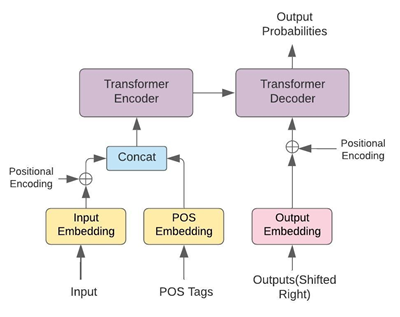}}
\caption{Transformer Architecture of POS Integration into the Positional Encoding}
\label{fig}
\end{figure}

\begin{table*}[t]
\caption{Corpus Statistics}
\begin{center}
\begin{tabular}{|c|c|c|c|c|c|}
\hline
\textbf{Dataset}&\textbf{Sentences}&\multicolumn{2}{|c|}{\textbf{Tokens}}&\multicolumn{2}{|c|}{\textbf{Unique Tokens}} \\
\cline{3-6} 
\textbf{} &\textbf{} & \textbf{\textit{English}}& \textbf{\textit{Sinhala}}& \textbf{\textit{English}}& \textbf{\textit{Sinhala}} \\
\hline
Training Set& 54914& 553006& 535091& 28742& 31524  \\
\hline
Validation Set& 1623& 23578& 22656& 4788& 5351  \\
\hline
Test Set& 1603& 19248& 18477& 4237& 4520  \\
\hline
Total& 58140& 595832& 577146& 37767& 41395  \\
\hline
\end{tabular}
\label{tab2}
\end{center}
\end{table*}

\begin{table}[t]
\caption{Best hyperparameter variations}
\begin{center}
\begin{tabular}{|l|c|}
\hline
\textbf{Encoder-Decoder Layers} & 5  \\
\hline
\textbf{Attention Heads} & 2  \\
\hline
\textbf{Input/Output Embedding Dimension} & 512  \\
\hline
\textbf{Embedding Dimension for FFN} & 2048  \\
\hline
\textbf{Label Smoothing} & 0.2  \\
\hline
\textbf{Dropout} & 0.4  \\
\hline
\end{tabular}
\label{tab3}
\end{center}
\end{table}

\begin{table}[t]
\caption{BLEU scores}
\begin{center}
\begin{tabular}{|p{6.2cm}|c|}
\hline
\textbf{Model}& \textbf{BLEU Score} \\
\hline
\textbf{Baseline Model} & 29.92  \\
\hline
\textbf{POS Integration with Input Embedding} & 30.84  \\
\hline
\textbf{POS Integration with Positional Encoding} & 28.75  \\
\hline
\end{tabular}
\label{tab3}
\end{center}
\end{table}

\subsection{Augmented Transformer Architecture}
The extracted linguistic information can be encoded in two ways in the Transformer architecture [4][14]. One method is to extract directly as a part of input embedding [4]. The other one is to treat this linguistic information as a type of positional information and thus adding them into the Transformer’s positional encoding unit [14]. We experimented with both methods in our research. Figure 1 and Figure 2 show the two augmented Transformer architectures for the two methods we used to incorporate POS information. 

The first model in Figure 1 takes the BPE encoded source sentence as the input and generates the original subword embedding and the POS embedding of every subword, then the model combines these two embeddings by concatenation. This concatenated embedding contains information about the subword as well as its context-based POS information. Then positional information about the subword is added at the positional encoding unit. Final resulting embedding is given to the Transformer encoder as the input. 

Figure 2 depicts the architecture of incorporating POS information into the positional encoding. As same as in the previous method, POS embeddings are generated for each subword. Here, the model concatenates the resulting POS embedding to the output embedding of the positional encoding unit. This concatenated embedding contains the positional information about that particular subword, in addition to the context-based POS information. This concatenated embedding is passed to the Transformer encoder.

\section{Experiments}

\subsection{Dataset and Preprocessing}

We focused on Sri Lankan official government document domain in our research to evaluate our models. A parallel corpus consisting of annual reports, committee reports, procurement, government acts and crawled content from government websites is used for the experiments. The parallel corpus is divided into training, validation and test sets which contain 54914, 1623 and 1603 sentence pairs respectively. Corpus statistics are shown in Table I.

Dataset was created by removing all single token sentence pairs which are number only, date only and uppercase reference only (eg: FTG, GSS). Considering the English dataset, sentences with 1-5 tokens were removed from the test dataset. The quality of the content has been ensured by professional human translators. Moses tokenizer [17] was used to tokenize the English sentences. A specifically built tokenizer [18] was used to tokenize the Sinhala sentences since existing tokenizers recognize a single Sinhala character as two characters.

We used BPE to segment the words in datasets into subwords. Since the models were trained at subword level, BPE was applied to all train, validation and test sets for source and target sides.

\subsection{Experimental Setup}

All train, validation and test sets were pos tagged. We used the Stanford POS tagger \footnote{https://nlp.stanford.edu/software/tagger.html} to get the POS tags for English corpus. The open-source Fairseq toolkit [19] was used for the experiments. The Transformer implementation was used as the basic architecture. To evaluate the systems, Bilingual Evaluation Understudy (BLEU) metric [20] was used.

\begin{table*}[t]
\caption{Examples from test set}
\centerline{{\includegraphics[width=17.5cm, height=3.5cm]{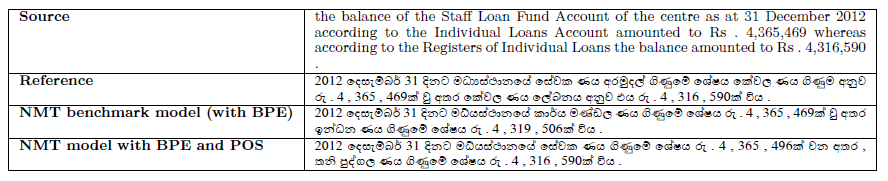}}}
\end{table*}

\subsection{Hyperparameters and Training}

We incorporated the hyperparameters of the Transformer benchmark model from [12] to train POS integrated models which is depicted in Table 2. We implemented various POS integrated models by changing the POS embedding dimension from 8 to 384. The total input embedding dimension is the concatenation of the subword embedding and the POS embedding. We kept this as a constant of 512 to ensure that the performance improvements are not due to the changing of the hyperparameters of the original benchmark model as proposed by Sennrich and Haddow [13]. Byte-Pair-Encoding was incorporated with all the benchmark and POS integrated models to overcome the out of vocabulary and rare word problems [21].

We chose Adam optimizer [22] with \({\beta}_{1}\) = 0.9, \({\beta}_{2}\) = 0.98 and \({\epsilon}\) = 10\(^{-8}\). The original learning rate proposed by Vaswani et al. [3] was used. We used beam search with a beam size of 5 and a length penalty of 1.2. The models were trained by batches. Each batch has 32 sentence pairs. Google Colab NVIDIA Tesla K80 GPU was used for model training.

\section{Results and Analysis}

To compare our POS integrated models, we used the Transformer BPE model [12] that resulted in a BLEU score of 29.92 as our baseline model. 

Using the baseline model, we experimented on two ways of integrating POS information as explained in the methodology section. However, the results obtained were completely different for the two models. For the first method, where the POS information was added to the input embedding, we were able to obtain a BLEU score of 30.84. Hence, our new model outperformed the benchmark model by a BLEU score of 0.92. 

However, our second model, where we incorporated POS information in the positional encoding, resulted in a BLEU score of 28.75, which is less than the BLEU score of the baseline model. We believe that this is due to the change of original positional information in the Transformer architecture.

Even though the BLEU gain from POS integration is not much significant, it is high compared to the existing linguistic knowledge incorporated research. The summary of the best BLEU scores obtained for the benchmark and POS integrated models are shown in Table III. Thus, it is clear that the translation quality can be improved by incorporating POS information.

Table IV depicts the sample translations obtained from the benchmark and the POS integrated models. The last row presents the translation obtained from the best POS integrated model (i.e. POS integration with Input Embedding) and it can be clearly seen that it has given a translation better than the baseline model similar to the expected Sinhala sentence.

\section{Conclusion}

The objective of this research is to implement an efficient domain-specific English to Sinhala NMT system using the Transformer architecture which incorporates POS information as an additional linguistic feature. This is the first NMT research done for English - Sinhala language pair incorporating POS information. We explored two ways of incorporating POS tags in the Transformer model. Our first approach of incorporating POS information into the input embedding shows impressive results compared to the benchmark Transformer model by increasing the quality of the translations and giving a BLEU score improvement of 0.92. Therefore, our research proves that linguistic knowledge indeed improves the quality of NMT models, as claimed in previous research [13]. However, since our second approach of incorporating POS information into the positional encoding resulted in a reduced BLEU score, further experiments should be carried out to find a way to obtain a high BLEU score for this. Hence, this can be considered as possible future work. We believe that this work will be beneficial for the low resource NMT research community to further enhance the quality of the low resource translation systems.

\section*{Acknowledgement}

We would like to thank the members of the National Languages Processing Centre at University of Moratuwa for developing the parallel corpus used in this research.

\end{document}